\documentclass[10pt,journal,compsoc]{IEEEtran}
\usepackage{cite}
\usepackage{multirow}
\usepackage{graphicx}
\usepackage{color}
\usepackage{dcolumn}
\usepackage{amsmath}
\usepackage{kantlipsum}
\usepackage{url}
\usepackage{microtype}
\usepackage{verbatim}
\usepackage{colortbl}
\usepackage{arydshln}
\usepackage{multirow}
\newcolumntype{K}[1]{>{\centering\arraybackslash}p{#1}}
\usepackage{booktabs}
\usepackage[flushleft]{threeparttable}
\usepackage{hhline}
\usepackage{algorithm}
\usepackage{algorithmic}
\usepackage{mathtools}

\ifCLASSOPTIONcompsoc
\usepackage[caption=false, font=normalsize, labelfont=sf, textfont=sf]{subfig}
\else
\usepackage[caption=false, font=footnotesize]{subfig}
\fi
\begin{document}
\title{TUTOR:  \textbf{T}raining Ne\textbf{u}ral Ne\textbf{t}works Using Decisi\textbf{o}n \textbf{R}ules as Model Priors}

\author{Shayan~Hassantabar,~Prerit~Terway,~and~Niraj~K.~Jha,~\IEEEmembership{Fellow,~IEEE}% <-this % stops a space
\thanks{
This work was supported by NSF under Grant No. CNS-1907381.  Shayan Hassantabar, Prerit Terway, and 
Niraj K. Jha are with the Department of Electrical \& Computer Engineering, Princeton University, 
Princeton, NJ 08544, USA, e-mail:\{seyedh, pterway,jha\}@princeton.edu.}}

\maketitle

\begin{abstract}
The human brain has the ability to carry out new tasks with limited experience. It utilizes prior learning experiences to adapt the solution strategy to new domains. On the other hand, deep neural 
networks (DNNs) generally need large amounts of data and computational resources for training. 
However, this requirement is not met in many settings.  To address these challenges, we propose the 
TUTOR DNN synthesis framework. TUTOR targets tabular datasets.  It synthesizes accurate 
DNN models with limited available data and reduced memory/computational requirements. 
It consists of three sequential steps. 
The first step involves generation, verification, and labeling of synthetic data. 
The synthetic data generation module targets both the categorical and continuous features. 
TUTOR generates the synthetic data from the same probability distribution as the real data.  It then 
verifies the integrity of the generated synthetic data using a semantic integrity classifier module.
It labels the synthetic data based on a set of rules extracted from the real dataset. 
Next, TUTOR uses two training schemes that combine synthetic and training data to learn the parameters 
of the DNN model.  These two schemes focus on two different ways in which synthetic data can be used 
to derive a prior on the model parameters and, hence, provide a better DNN initialization for 
training with real data.  In the third step, TUTOR employs a grow-and-prune
synthesis paradigm to learn both the weights and the architecture of the DNN to reduce model size while 
ensuring its accuracy.  We evaluate the performance of TUTOR on nine datasets of various sizes.
We show that in comparison to fully-connected DNNs, TUTOR, on an 
average, reduces the need for data by $5.9\times$ (geometric mean), improves accuracy by $3.4\%$, and reduces the number of parameters (floating-point operations) by $4.7\times$ 
($4.3\times$) (geometric mean). 
Thus, TUTOR enables a less data-hungry, more accurate, and more compact DNN synthesis. 
\end{abstract}

% \begin{IEEEkeywords}
% Architecture synthesis; compact network; compression; deep neural network; synthetic data generation.
% \end{IEEEkeywords}

\IEEEpeerreviewmaketitle

% \ifCLASSOPTIONcompsoc
% \IEEEraisesectionheading{\section{Introduction}\label{sec:introduction}}
% \else
% \section{Introduction}
% \label{sec:introduction}
% \fi

\section{Introduction}
\label{sec:introduction}

\IEEEPARstart{A}{n} exceptional aspect of human intelligence is its ability to leverage prior experiences to solve problems in a new domain with limited experience.  A similar ability 
to learn efficiently under limited available data and computational resources is 
also a desirable attribute of artificial intelligent (AI) agents.

Over the last decade, deep neural networks (DNNs) have revolutionized a variety of application domains,
such as computer vision, speech recognition, and robotic control. However, data scarcity is still a 
limiting factor in training DNNs for various applications, such as healthcare and cognitive
modeling of human decisions \cite{hassantabar2020coviddeep, hassantabar2021mhdeep, bourgin2019cognitive}.  Many such
applications offer just a few thousand data instances for training.  Training of DNN models with such small datasets may lead to overfitting, hence limit their applicability to real-world environments.

Some of the drawbacks of the current DNN training process are as follows:

\begin{itemize}
    \item \emph{Need to obtain and label large datasets}: Collecting a large number of data
    instances and manually labeling them is costly and time-consuming, even more so in domains where experts are needed to label the data instances. As a result, reducing the cost of labeling is an active area of research \cite{enayati2021visualization}.
    \item \emph{Ignoring domain knowledge}: Training DNN architectures only on available training data does not take into account available expert domain knowledge or the set of rules that may have been derived for the domain. As a result, the process is not efficient and, in turn,
    exacerbates the need for large datasets.
    \item \emph{Substantial redundancy}: Most DNN architectures need substantial storage, memory bandwidth, and computational resources for training. However, recent work shows that it is
possible to significantly reduce the number of parameters and floating-point operations (FLOPs), with no loss in accuracy ~\cite{han2015learning, dai2019nest}.
\end{itemize}{}

There has been a recent spurt of interest in combining domain-specific information in the form of
rule-based AI with modern statistical AI~\cite{richardson2006markov, serafini2016logic}. In such
works, rule-based AI can act as an inductive bias to enhance the learning efficacy of novel tasks. 
Imposing prior knowledge mitigates the need for a large amount of training data. 
However, extracting domain knowledge or creating a set of rules for a novel domain is often 
challenging as it requires specific expertise in the area.  Nevertheless, domains, such as natural 
language processing, often have pre-defined rules that make such a combination attractive
\cite{balestriero2017neural}. 

Researchers have also explored various synthetic data generation methods to address a number of 
problems, such as preserving privacy when sharing data, testing new tools, and increasing dataset size. 
Generative models, such as variational auto-encoders \cite{pu2016variational} and 
generative adversarial networks (GANs) \cite{goodfellow2014generative}, provide an appealing solution 
for generating well-performing synthetic data. 
Using image and text data, these models learn the probability distribution of training data. 
Recently, GANs have also been used in the generation of tabular synthetic data \cite{park2018data}.  
They have been shown to generate synthetic data that are close to the real data distribution. 
However, the use of neural network models with a large number of parameters in the generator and 
discriminator components of GANs necessitates large training datasets to generate high-quality 
synthetic data.  In this work, we assume we have access to only a limited number of data instances 
in order to synthesize accurate models that generalize well.  As a result,  we aim to develop a 
synthetic data generation module that is sample-efficient and maximizes the use of the small available 
data. 

To address the above problems, we propose a new DNN training and synthesis framework called
TUTOR, which is particularly suitable for tasks where limited training data and computational resources are 
available.
It relies on the generation of synthetic data from the same probability 
distribution as the available training data, to enable the DNN to start its training from a better 
initialization point.  
TUTOR consists of three sequential stages: (1) synthetic data generation, verification, and 
labeling, (2) training schemes for incorporating synthetic and real data into the training process, and 
(3) grow-and-prune DNN synthesis to ensure model compactness as well as to enhance performance.

Data augmentation is an effective technique, often used in image classification applications, to 
increase dataset size, help the model learn invariance, and regularize the model 
\cite{lingchen2020uniformaugment}.  However, TUTOR targets tabular datasets.  
It generates synthetic data by modeling the joint multivariate distribution of the given dataset and 
sampling from the learned distribution. 
The synthetic data generation module simultaneously generates categorical and continuous features.
To ensure the semantic integrity of synthetic data, TUTOR uses a semantic integrity classifier module 
that verifies the validity of the generated synthetic data. This classifier verifies the values of 
categorical features with respect to other features of the synthetic data instance.  We label the 
synthetic data by incorporating a set of rules extracted from real data. These rules are obtained 
using a random forest model trained on real data.  Finally, TUTOR uses the synthetic data alongside the 
real data in two training schemes.  The intuition behind using synthetic data in DNN training is to 
provide a suitable inductive bias to the DNN weights.  It thus mitigates the problem of limited 
data availability.

To address the problem of network redundancy, we use the grow-and-prune synthesis 
paradigm \cite{dai2019nest} by leveraging three different operations: connection growth, neuron 
growth, and connection pruning.  It allows the DNN architecture to grow neurons and connections to 
adapt to the prediction problem at hand. Subsequently, it prunes away the neurons and connections of 
little importance. In addition, it does not fix the number of layers in the architecture
beforehand, enabling the architecture to be learned during the training process.

The major contributions of this article are as follows:

\begin{itemize}
    \item TUTOR addresses two challenges in the design of predictive DNN models: lack of
    sufficient data and computational resources.  It focuses on tabular datasets and addresses the 
    lack of sufficient data by generating synthetic data from the same probability distribution as the 
    training data using three different density estimation methods. 
    It uses semantic integrity classifiers to verify the integrity of the generated synthetic data. 
    By relying on decision rules generated from a random forest machine learning model, it obviates the 
    need for domain experts to label the synthetic data. 
    \item TUTOR uses a training flow to combine synthetic and real data to train the DNN models and 
    learn both the weights and the architecture by using a grow-and-prune synthesis paradigm.  
    This addresses the problem of fixed network capacity while reducing memory and computational costs.
    \item Less data: TUTOR requires $5.9\times$ fewer data instances to match or exceed the 
    classification accuracy of conventional fully-connected (FC) DNNs.
    \item Efficiency: TUTOR uses $4.7 \times$ fewer parameters and 4.3$\times$ fewer 
    FLOPs relative to FC DNNs.
    \item Accuracy: TUTOR enhances classification accuracy by $3.4\%$ over conventional FC DNNs.
    \item Privacy enhancement: The synthetic data generated by TUTOR can be separately used to train 
    a DNN, with little to no drop in accuracy, without the need for real data.  This is useful in 
    applications where privacy issues make sharing of real data infeasible.
\end{itemize}

The rest of the article is organized as follows. Section \ref{sect:related} covers related
work. Section \ref{sect:background} provides some necessary background and the
theoretical motivation behind this work. Then, in Section \ref{sect:methodology}, we discuss
the proposed TUTOR framework in detail. Section \ref{sect:results} presents evaluation results
for TUTOR. In Section \ref{sect:discussion}, we provide a short discussion on analogies between the 
human brain and the TUTOR training flow. Finally, Section \ref{sect:conclusion} concludes the paper. 

\section{Related Work}
\label{sect:related}
In this section, we review several related works along four different dimensions. First, we discuss how logical AI can be combined with data-driven statistical AI. Second, we review work that combines decision trees with DNNs. Third, we 
discuss some of the recent work in efficient DNN synthesis.  
Finally, we discuss various data augmentation and synthetic data generation methods. 

\subsection{Combining Logical AI with Statistical AI}
Recently, there have been several attempts to combine propositional and first-order logic (FOL) with 
the statistical techniques of machine learning. 
For instance, Markov logic network (MLN) 
\cite{richardson2006markov} is based on probabilistic logic that combines Markov networks with FOL. 
The first-order knowledge base (KB) is viewed as imposing a set of hard constraints, where violation of a formula reduces the probability of the targeted application world to zero. An MLN softens 
these constraints by associating a weight with each FOL formula in the KB. These weights are 
proportional to the probability of the FOL formula being true.  The incorporation of FOL enables 
reasoning about pre-existing domain knowledge. 
As a result, the Markov networks model uncertainty in reasoning. 
However, the main drawback of 
an MLN is the need for domain expertise to obtain the formulas, thereby limiting its usage.  

% Nodes in the MLN define the FOL predicates. An arc exists between two nodes if the predicates appear in the 
% formula. The probability of a world $x$ being true is given by:
% $$
% P(x)=\frac{1}{Z} \exp \left(\sum_{k \in \text {ground formulas}} w_{k} f_{k}(x)\right)
% $$

% \noindent
% where $Z$ is a normalization constant and $w_{k}$ is the weight of the $k^{th}$ formula given by $f_k(x)$. A formula 
% in which the variables are replaced by constants is referred to as a ground formula. 
% MLN learns the weights by maximizing the likelihood of a query given some evidence. Prior knowledge imposes 
% an inductive bias that allows learning with small amounts of data.  

The logic tensor network (LTN) \cite{serafini2016logic} uses fuzzy logic to combine DNNs with FOL. 
Logical predicates represent concepts that are combined to form logical formulas.  Fuzzy logic 
enables a formula to be partially true and take a value between 0 and 1.  In LTN, a DNN learns the 
membership of an object in a concept.  Each concept or formula is a point in a feature space and the DNN outputs a number between 0 and 1 to denote the degree of membership. 

\subsection{Combining Decision Trees with DNNs}
There are several works that combine DNNs with decision trees. 
Neural-backed decision trees (NBDTs) \cite{wan2020nbdt} combine decision trees with a DNN to
enhance DNN explainability.  
% Decision trees make higher-level decisions and the DNN makes low-level 
% decisions.  NBDTs also make intermediate decisions as opposed to the DNN that outputs decisions only 
% at the last layer.  
NBDTs achieve performance close to that of the DNN while improving explainability. 
A neural decision tree \cite{balestriero2017neural} incorporates a nonlinear splitting criterion
in its nodes by using a DNN to make the splitting decision. Deep neural decision forests 
\cite{kontschieder2015deep} present an end-to-end learning framework based on convolutional neural 
networks (CNNs) and a stochastic differentiable decision tree for image classification applications.
This approach uses back-propagation to learn the parameter to split at a node.
Deep neural decision trees
\cite{yang2018deep} use a soft binning function modeled by a DNN to split nodes into multiple leaves. 

Initializing the weights of a DNN from a good starting point can also be used to improve its performance. Humbird et al.  \cite{humbird2018deep} use a decision tree model to construct the architecture and initialize the weights of the DNN. 
The DNN architecture is obtained using the decision paths of a decision tree.
The tree acts as a warm-start for the training process.

\subsection{Efficient DNN Synthesis}
To reduce the need for data, a popular approach involves the use of transfer learning 
\cite{oquab2014learning} to transfer weights learned for one task to initialize weights of a network for a similar task. This approach is based on the fact that some of the learned features are similar across related tasks.

A common approach for reducing computational cost is the use of efficient building blocks. For 
example, MobileNetV$2$ \cite{sandler2018mobilenetv2} uses inverted residual blocks to reduce model 
size and FLOPs.  ShuffleNet-v$2$ \cite{ma2018shufflenet} uses depth-wise separable 
convolutions and channel-shuffling operations to ensure model compactness. Spatial convolution is one 
of the most expensive operations in CNN architectures. To reduce its computational cost, 
Shift \cite{wu2018shift} uses shift-based modules that combine shifts and point-wise convolutions 
that significantly reduce computational cost and storage needs.  FBNet-v$2$ uses differentiable neural 
architecture search to automatically generate compact architectures. Efficient performance
predictors, e.g., for accuracy, latency, and energy, are also used to accelerate the DNN search 
process \cite{dai2018chamnet, hassantabar2019steerage}. 

DNN compression methods can be used to remove redundancy in DNN models. Han et 
al.~\cite{han2015deep} proposed a pruning methodology to remove redundancy from large CNN 
architectures, such as AlexNet and VGG. Pruning methods are also effective on recurrent neural 
networks \cite{han2017ese}. Combining network growth with pruning enables a sparser, yet more 
accurate, architecture. Dai et al.~\cite{dai2019nest, dai2018grow} use the grow-and-prune 
synthesis paradigm to generate efficient CNNs and long short-term memories (LSTMs). 
SCANN \cite{hassantabar2019scann} uses feature dimensionality reduction alongside grow-and-prune 
synthesis to generate very compact models for deployment on edge devices and Internet-of-Things 
(IoT) sensors.

Orthogonal to the above works, low-bit quantization of DNN weights can also be used to reduce 
FLOPs. A ternary weight representation is used in \cite{zhu2016trained} to significantly reduce 
computation and memory costs of ResNet-$56$, with a limited reduction in accuracy. 

\subsection{Data Augmentation and Synthetic Data Generation}
Learning with sparse data is challenging as machine learning algorithms, especially DNNs, face the 
overfitting problem in a small-data regime.  To overcome this problem, several techniques have been 
developed to augment the dataset. Data augmentation for images enhances 
the accuracy of image classification and reinforcement learning tasks.  AutoAugment 
\cite{cubuk2018autoaugment} uses reinforcement learning to obtain a policy for optimal image 
transformation to improve performance on various computer vision tasks. 
The optimal transformation is dependent on the dataset.  UniformAugment \cite{lingchen2020uniformaugment} augments images by uniformly 
sampling from the continuous space of augmentation transformations, hence, avoiding the costly search 
process for finding augmentation policies.  Laskin et al. \cite{laskin2020reinforcement} use two new 
augmentation schemes based on random translation and random amplitude scaling alongside standard 
augmentation techniques (crop, cutout, flip, rotate, etc.). They use these augmentation schemes to 
achieve a state-of-the-art performance for reinforcement learning tasks. Antoniou et 
al.  \cite{antoniou2017data} use GANs to augment the images for few-shot learning in the low-data 
regime tasks. 

Privacy is an important concern when sharing data with partners or releasing data to the public. 
As a result, researchers have investigated techniques to generate synthetic data with performance 
similar to that of real data.  TableGAN \cite{park2018data} is a GAN-based approach for creating 
synthetic tabular data. It uses a GAN architecture with three components: generator, discriminator, and 
a classifier that learns correlations between the labels and other attributes of the table. 
Synthetic data generated by TableGAN is shown to achieve high model capacity while offering
protection against membership attacks. 
TGAN \cite{xu2018synthesizing} generates tabular data based on GANs and uses an LSTM 
with attention mechanisms to generate the synthetic data column by column. 
CTGAN \cite{xu2019modeling} improves tabular synthetic data generation by introducing conditional 
vectors for categorical features and training by sampling the different conditional vectors. 

\section{Background}
\label{sect:background}

In this section, we discuss background material that motivates the TUTOR framework. First,
we discuss different probability density estimation methods. We use such methods in the
TUTOR synthetic data generation module. Next, we discuss the Hierarchical Bayes and Empirical Bayes
concepts. We also discuss linking of gradient-based learning with hierarchical Bayes. 
These methods motivate our training schemes (explained in Section \ref{sect:schemes}) in the way 
we combine synthetic and real data to learn the DNN model parameters. 

\subsection{Probability Density Estimation}
\label{sect:PDE}
The probability density function (pdf) is a fundamental concept in statistics. It is used to compute 
various probabilities associated with a random variable.  Probability density estimation deals with 
estimating the density of function $f$ with $\hat{f}$, given random samples 
$x_1, \dots, x_n \stackrel{}{\sim} f$. 

Predominantly, probability density estimation can be categorized into two groups: parametric and 
non-parametric. In parametric density estimation, pdf $f$ is assumed to be a member of a parametric 
family. Hence, density estimation is transformed into finding estimates of the various 
parameters of the parametric family. For example, in the case of a normal distribution, density 
estimation aims to find the mean $\mu$ and standard deviation $\sigma$. In general, the classic 
parametric mixture of $C$ Gaussian models has a pdf of the form:

$$
p(x_{i}|\Theta) = \Sigma_{c=1}^{C}  p(x_{i}|z_{i}=c, \Theta) p(z_{i}=c|\Theta)
$$
where each component $c$ is a $d$-dimensional  multivariate Gaussian distribution with mean vector ${\mu}_{c}$ and covariance matrix  $\Sigma_{c}$ in the form of:
\begin{equation*}
\begin{split}
   p(x_{i}|z_{i}=c, \Theta) = \left(\frac{1}{2\pi}\right)^{(d/2)} |\Sigma_{c}|^{-1/2} \\
   \text{exp} \left( -\frac{1}{2} (x_{i}-\mu_{c})^{T} \Sigma_{c}^{-1} (x_{i}-\mu_{c}) \right) 
\end{split}
\end{equation*}
where $|\Sigma_{c}|$ is the determinant of the covariance matrix. In addition, the mixture probabilities are modeled as categorical:
$$
p(z_{i} = c|\Theta) = \theta_{c}
$$

A non-parametric density estimation scheme can generate any possible pdf. It does not make any 
distributional assumptions and covers a broad class of functions. Kernel density estimation is a 
non-parametric density estimation approach that approximates a probability distribution as a sum of 
many \emph{kernel} functions. Each kernel function $K$ satisfies the following property:

$$
\int_{-\infty}^{\infty} K(y) dy = 1 \quad , \quad K(y) > 0 \quad \forall {y}
$$

There may be various options for choosing function $K$, such as the pdf of the normal distribution. 
Another important parameter is the kernel \emph{bandwidth}, $h$, that rescales the kernel
function.  Kernel density estimation can be formulated as:

$$
\hat{f}(y) = \frac{1}{nh} \sum_{i=1}^{n} K(\frac{y-x_{i}}{h})
$$
where $n$ is the number of samples and $x_i$ is the $i^{th}$ sample. 

\subsection{Hierarchical Bayes and Empirical Bayes}
\label{sect:bayes}

Creating sub-models can be a way to model the distribution of data with a hierarchical structure. 
Each sub-model has its own parameters that are interlinked to model the complete distribution. 
Hierarchical Bayes models the data distribution at a prior level by using Bayes inference that links the parameters and propagates uncertainties among the sub-models. 
It uses the Bayesian network assumption: variables are conditionally independent of their non-descendants, given their parents.
The equation below shows hierarchical Bayes modeling of the distribution of data $X$ 
\cite{malinverno2004expanded}.

$$
X \longrightarrow \theta \longrightarrow \theta_{1} \longrightarrow \theta_{2} \longrightarrow 
\ldots \longrightarrow \theta_{t}
$$

\noindent
where $\theta$ is the parameter and $\theta_{1}, \theta_{2}, \ldots , \theta_{t}$ are the 
hyperparameters. 
Based on the Bayes assumption, using only the 
following conditionals specifies the complete model: $[X|\theta]$, $[\theta|\theta_{1}]$, $[\theta_{1}|\theta_{2}]$, $\ldots$, $[\theta_{t-1}|\theta_{t}]$.
\noindent
By exploiting the inherent structure of the distribution, hierarchical modeling reduces the number of parameters needed in its modeling. Hence, it is less prone to overfitting. 
Empirical Bayes \cite{malinverno2004expanded} is similar to Hierarchical Bayes. 
However, Empirical Bayes determines the hyperparameters by using the data instead of the hyper-prior distribution. 
% In addition, estimation is done through non-Bayesian techniques like maximum-likelihood or method of 
% moments applied to the validation data.  
% These hyperparameters are used for inference on new data
% instances. 

The two training schemes of TUTOR (explained in Section \ref{sect:schemes}) model DNN parameter learning
with Hierarchical Bayes.  In addition, similar to Empirical Bayes, TUTOR 
estimates the hyperparameters by using the available data.  In this process, it first uses probability density estimation to estimate the distribution of the training dataset. 
It uses the sum of the log-probabilities of the data instances in the validation set to determine the 
hyperparameter $\lambda$ of the probability density estimation function.  
It generates synthetic data by sampling from this function.
In the training schemes, it learns parameters $\theta$ of the DNN model by using the synthetic data to pre-train the weights of the model.  Therefore, we can obtain a Hierarchical Bayes model 
of the parameters as follows:
$$
X \longrightarrow \lambda \longrightarrow \theta 
$$

\noindent
The posterior distribution can be written as:

$$p(\lambda, \theta \mid X) \propto p(X \mid \theta, \lambda) p(\theta \mid \lambda) p(\lambda)$$ 

\noindent
where $X$ denotes the combined training and validation set,  $\theta$ the parameter of the model 
after training with synthetic data, and $\lambda$ the hyperparameter of the probability density 
estimation function.

\subsection{Linking Gradient-based Learning with Synthetic Data and Hierarchical Bayes}
\label{sect:gradient+bayes}

Gradient-based learning combined with a Bayesian network enables the transfer of knowledge across various domains in a sample-efficient manner. Model-agnostic meta-learning (MAML) 
\cite{finn2017modelagnostic} trains a model across a variety of tasks and fine-tunes it using 
gradient descent on a new learning task.  
As a result, MAML uses other related tasks to initialize DNN weights instead of 
using random initialization.  Grant et al. \cite{grant2018recasting} cast MAML as Empirical Bayes 
using a hierarchical probabilistic model. 
It learns a prior that is adapted to new tasks to augment gradient-based meta-learning. 
Learning on one task influences the parameters of another task through a meta-level parameter that 
determines the task-specific parameters.  

In a similar fashion to MAML, TUTOR generates synthetic data and uses them to learn parameters 
$\theta$ of the model. Then, it learns parameters $\phi$ of the DNN from the initialization with 
$\theta$, followed by gradient descent using real training data.

%$$
%\phi = \theta - \alpha \nabla_{\theta}\mathcal{L}\left(f_{\theta}\right)
%$$

\section{Synthesis Methodology}
\label{sect:methodology}
In this section, we discuss the TUTOR DNN synthesis framework in detail. 
First, we give a high-level overview of its DNN training methodology. 
We then zoom into each part of the methodology.
First, we explain the three synthetic data generation methods, the semantic integrity classifier module, and the synthetic data labeling method. 
Next, we focus on the two training schemes that combine the real and 
synthetic data in the training process.  
We then explain our grow-and-prune synthesis algorithm.  Finally, we discuss
the use of synthetic data alone as a proxy for real data to train DNNs, with a little to no loss in
accuracy, for the sake of privacy enhancement.

\subsection{Framework Overview}
\begin{figure*}[!hbt]
    \centering
    \includegraphics[scale=0.5]{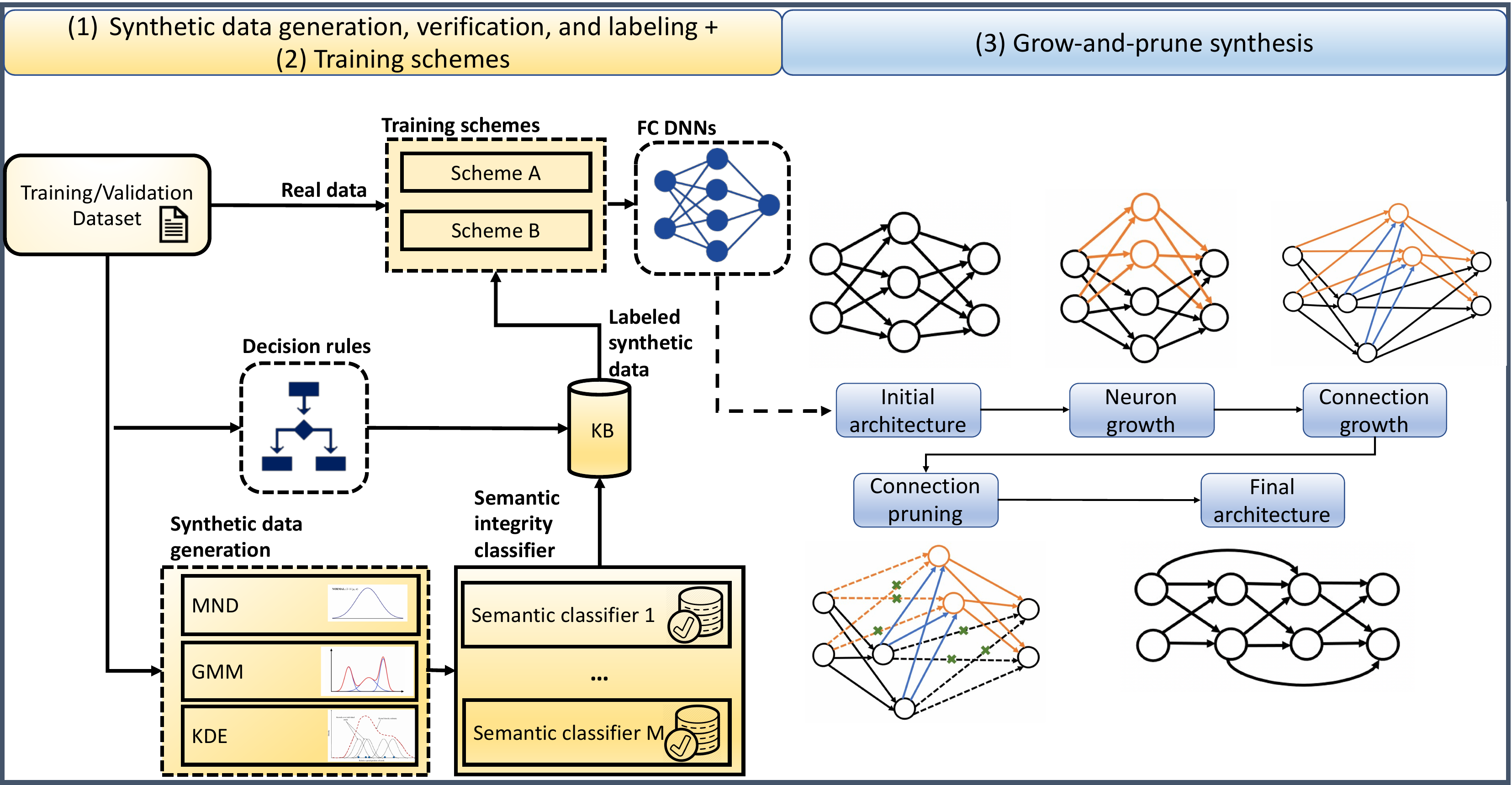}
    \caption{Block diagram of the TUTOR DNN synthesis framework: (1) synthetic data generation, 
verification, and labeling, (2) two training schemes, and (3) grow-and-prune synthesis.}
\label{fig:diagram}
\end{figure*}

We illustrate the proposed TUTOR framework in Fig.~\ref{fig:diagram}.
It consists of three main parts: (1) synthetic data generation, verification, and labeling with decision rules, 
(2) two training schemes that combine synthetic data with real data, and (3) grow-and-prune 
synthesis to decrease network redundancy and computational cost for inference while improving 
performance. 

The synthetic data generation module uses the training and validation sets as inputs to generate 
synthetic data instances.  It targets both continuous and categorical features of the tabular data. 
We formulate this step as an optimization problem. 
The objective is to find the best set 
of hyperparameters ($\lambda$) of the probability density estimation function computed on the 
training data ($\hat{f}(X_{train}|\lambda)$). 
To this end, we use the sum of the log probabilities of instances in the validation set
($\texttt{score}(X_{validation})$) as a measure for comparing different functions.
We sample a pre-defined number of instances ($count$) from the density estimation function $\hat{f}$
with an optimal set of hyperparameters $\lambda^{*}$ to generate synthetic data:
$$
\lambda^{*} = \underset{\lambda}{\arg\max} \left( \hat{f}(X_{train}|\lambda).\texttt{score}(X_{validation}) \right)
$$
$$
X_{syn} = \hat{f}(X_{train}|\lambda^{*}).\texttt{sample}(count)
$$
To synthesize the correct values for categorical features, we treat them differently from the continuous features, using the one-hot encoding technique.  
To ensure the validity of the categorical features relative to other data features, we use several 
semantic integrity classifiers to verify the generated synthetic data. 
Section~\ref{sect:syn-data} explains the process of synthetic data generation and verification.

To label the synthetic dataset, we use decision rules to obviate the need for the expensive and
laborious process of labeling by an expert.  To do so, we use a random forest model trained on data 
used as a KB. Section~\ref{sect:labeling} explains this process.

In the two training schemes, A and B, TUTOR uses synthetic data alongside the original training and 
validation data.  As explained in Section \ref{sect:gradient+bayes}, these schemes learn the DNN 
weights by combining gradient-based learning with Hierarchical Bayes modeling.  TUTOR uses synthetic 
data to learn parameters $\theta$ of the DNN model that enable better initialization for the next 
training step.  Then, it learns parameters $\phi$ of the DNN by performing gradient-descent training 
with real training data, starting from initialization point $\theta$. Section~\ref{sect:schemes} 
explains these two training schemes.

Finally, we perform grow-and-prune synthesis \cite{dai2019nest, hassantabar2019scann} on the 
FC DNN models that are the outputs of Schemes A and B.  This step uses gradient-based learning to 
obtain both the final architecture and the final DNN parameters.  Grow-and-prune synthesis 
uses three different architecture-changing operations that are explained in Section~\ref{sect:gp}. 

\subsection{Synthetic Data Generation}
\label{sect:syn-data}
In this section, we discuss how TUTOR models tabular data distribution and sampled rows from the 
distribution to generate synthetic data.  Tabular data $X$ contains $N$ continuous features and 
$M$ categorical features. Each feature is assumed to be a random variable.  The goal is to model 
the joint distribution of all the features, given the observations provided by real data instances. 
To tackle this problem, TUTOR uses both parametric and non-parametric density estimation methods. To 
deal with categorical data columns in the table, it uses one-hot encoding to model the value of each 
categorical feature for each data instance.  It also verifies the values of these categorical features
relative to other synthetic data features.  These values should be semantically correct.
For example, if a continuous feature representing blood glucose level has a value of 120 mg/dL, a 
categorical feature showing whether or not the subject has diabetes cannot have a value of 1 for
the following reason: the blood glucose level is too low for the subject to be diabetic. 
We introduce a module called the semantic integrity classifier that verifies that the generated 
synthetic data are semantically correct.  We then use supervised machine learning models to label 
verified synthetic data. 

\subsubsection{Encoding the categorical features}
We use the one-hot encoding to encode categorical data. Hence, for categorical feature $F$ with 
$k$ different values, we use $k$ columns to represent feature $F$, with $1$ in the column 
corresponding to its value and 0 in the other columns. In addition, after generating synthetic data, 
we use the $softmax$ operator on the synthetic values corresponding to $k$ possible different values of the feature to specify the value of that feature for the synthetic data instance. 
Note that categorical features with a very high cardinality in the presence of a low number of data 
instances may lead to incorrect values for these features in the synthetic data. Hence, using a 
semantic integrity classifier to check for the validity of these values is important. We explain this 
part in more detail in Section~\ref{sect:semantic-classifier}.  

\subsubsection{Probability density estimation}
In this section, we discuss the three probability density estimation methods we use to model the 
distribution of tabular data.  The first approach involves the use of 
multivariate normal distribution to estimate the probability distribution of training data.
This is an example of parametric density estimation.  We also use an approach based on kernel density 
the estimation that resides on the other side of the spectrum, being a non-parametric density 
estimation approach.  Finally, we use Gaussian mixture models (GMMs) to estimate the training data 
distribution. The GMM approach can be considered to effect a trade-off between the other two 
approaches. It is based on the sum of a limited number of multivariate normal distributions. 

\noindent
\textbf{Multivariate normal distribution (MND):}
Next, we discuss modeling data with a multivariate normal distribution. 
The parameters of this distribution are computed based on the training data. 
The sample mean vector $\hat{\mu}$ can be estimated as:
$$
\hat{\mu} = \frac{1}{N} \sum_{i=1}^{N} x_{i}
$$
where $N$ is the total number of training data instances and $x_i$ is the $d$-dimensional vector 
that depicts the $i^{th}$ training instance. 

To estimate the $d \times d$ covariance matrix of the multivariate normal distribution, we use the 
maximum likelihood estimator of the covariance matrix:
$$
\hat{\Sigma} = \frac{1}{N} \sum_{i=1}^{N} (x_{i} - \hat{\mu})(x_{i} - \hat{\mu})^{T}
$$

Thus, a multivariate normal distribution with a $d$-dimensional mean vector 
$\hat{\mu}$ and $d \times d$ covariance matrix $\hat{\Sigma}$ estimates the density. 
The joint density function of a random vector $X$ based on this density function is given by:

$$
{\textstyle
p_{X}(x|\hat{\mu}, \hat{\Sigma}) = \left( \frac{1}{2\pi} \right) ^ {d/2} |\hat{\Sigma}|^{-1/2} \text{exp} \left( -\frac{1}{2} (x-\hat{\mu})^{T} \hat{\Sigma}^{-1} (x-\hat{\mu}) \right)
}
$$

\noindent
where $|\hat{\Sigma}|$ is the determinant of matrix $\hat{\Sigma}$. 
When the number of available data points is small, we can constrain the covariance matrix to be 
diagonal to reduce the number of independent parameters in the model.  The synthetic data are
generated by sampling the following distribution:
$$
X \sim \mathcal{N} (\hat{\mu}, \hat{\Sigma})
$$

\noindent
\textbf{Gaussian mixture model (GMM):}
\label{sect:GMM}
The second approach involves the use of a multi-dimensional GMM to model the training data 
distribution.  The data are modeled as a mixture of $C$ Gaussian models that lead to a pdf of the 
form:
$$
p_{X}(x|\Theta) = \Sigma_{c=1}^{C}
p_{X|Z}(x|z=c, \Theta)
p_{Z}(z=c|\Theta)  
$$
where $\Theta$ depicts the parameters of the Gaussian model, $X$ the observed variables, and $Z$ 
the hidden state variables that indicate Gaussian model assignment, with the prior probability of 
each model $c$ being:  
$$
p_{Z}(z=c) = \theta_{c}
$$
Hence, the GMM can simply be written as:

$$
p_{X|C}(x)=\sum_{c=1}^{C}  \mathcal{N}\left(x|\mu_{c},\Sigma_{c}\right)
\pi_{c}
$$
Here, ${\mu}_{c}$ represents the mean vector, $\Sigma_{c}$ the covariance matrix, and $\pi_{c}$ 
the weight of GMM component $c$, with a total of $C$ components.

The drawback of this approach lies in its learning complexity.  To address this issue,
we use the iterative Expectation-Maximization ($EM$) algorithm to determine the GMM parameters. 
The EM algorithm solves this optimization problem in two steps:

\begin{itemize}
    \item E-step: given the current parameters $\theta_{i}$ and observations in data $\mathcal{D}$, 
estimate the values of $z_{i}$.
    \item M-step: with the new estimate of $z_{i}$, find the new parameters, $\theta_{i+1}$.
\end{itemize}

In the $E$-$step$, for each training data instance $x_{i}$, the most likely cluster 
$z_{i}$ is obtained. In the $M$-$step$, the cluster parameters, i.e., cluster mean $\mu_{c}$ and 
covariance matrix $\Sigma_{c}$, are obtained from the current cluster assignments.  This procedure is 
repeated until convergence.  Several choices are available for cluster shape \textit{(diagonal, 
spherical, tied, full)} that can be used to control the degree of cluster freedom. 
The choices span the spectrum from the most general to more specific mixture models.
Although the \textit{full} type generates a more general mixture model, it also increases the number 
of independent model parameters.  Hence, with limited available data, using a more specific model 
may potentially improve performance by reducing the number of independent model parameters. 
Hence, we explore these four choices in our setup.

To avoid overfitting on the training data, the likelihood of validation data is obtained by varying
the number of mixtures, $C$. The sum of log probabilities of instances in the validation set is used 
to evaluate the quality of the model. The number of components that maximizes this criterion is 
chosen as the optimal number of components required to model the distribution: 
$$
C^{*} = \underset{C}{\arg\max} \left( p_{X|C}(x).\texttt{score}(X_{validation}) \right)
$$
The GMM parameters thus obtained are then used to generate synthetic data by sampling a pre-defined 
number of instances ($count$) from this distribution.
$$
X_{syn} = p_{X|C^{*}}(x).\texttt{sample}(count)
$$

\noindent
\textbf{Kernel density estimation (KDE):}
\label{sect:kde}
As opposed to the parametric mixture models, non-parametric mixture models assign one component to 
every training example. In general, this can be formulated as:
$$
p(x|\Theta) = \sum_{i=1}^{N} p_{Z}(z_{i}=c|\Theta)p_{X|Z}(x|z=c, \Theta)
$$
In our setup, we use the pdf of the normal distribution as the kernel function.  An important hyperparameter of the KDE model is kernel bandwidth $h$ that controls the smoothness of the estimated function.  The bandwidth has a large impact on the bias-variance trade-off of the estimator. 
Essentially, high-variance models estimate the training data well, but suffer from overfitting on the 
noisy training data. On the other hand, high-bias estimators are simpler models, but suffer from 
underfitting on the training data and fail to capture important regularities. In the KDE models, 
small bias is achieved when $h \rightarrow 0$ and small variance when $h \rightarrow \infty$. 
Therefore, the choice of the $h$ value directly impacts function performance on unseen data. 
Hence, as in the case of finding the optimal number of components in a GMM model, we use a
validation set to find the optimal bandwidth value $h^{*}$. We use the sum of log probabilities of 
the instances in the validation set to evaluate the KDE probability estimator: 
$$
h^{*} = \underset{h}{\arg\max} \left( p_{X|h}(x).\texttt{score}(X_{validation}) \right)
$$

After obtaining all the parameters of the KDE estimator, we sample a pre-defined number of 
instances ($count$) from this distribution to generate the synthetic data:

$$
X_{syn} = p_{X|h^{*}}(x).\texttt{sample}(count)
$$

\subsubsection{Semantic integrity classifier}
\label{sect:semantic-classifier}
To ensure the quality of the synthetic data, and verify the semantic integrity of the categorical 
feature values, we utilize the semantic integrity classifier module. 
Given a synthetic record, the task of the semantic integrity classifier is to predict the values of 
the categorical features based on the values of other continuous features in that data instance. 
We train a semantic integrity classifier for each categorical feature.
Algorithm~\ref{alg:semanctic-integrity} summarizes the semantic integrity verification process for one 
categorical feature.  In the case of a mismatch between the value predicted by the semantic integrity 
classifier and the value of the categorical feature in the synthetic data, we discard the data 
instance.  The classifier model can be chosen from among many options.  In our implementation, we use 
random forest models to implement the semantic integrity classifiers. We use validation accuracy to 
optimize the hyperparameters of the model. 

\begin{algorithm}[h]
    \caption{Semantic integrity classifier}
    \label{alg:semanctic-integrity}
    \begin{algorithmic}[l]
        \REQUIRE \textit{($X$)}: Training data; \textit{($X_{syn}$)}: generated synthetic data;
\textit{$F_{cont}$}: the set of continuous features; \textit{$F_{cat}$}: the categorical feature;
        \textit{$CLF$}: classifier model
        
        \STATE Train \textit{$CLF$} on $(X[:, F_{cont}] , X[:, F_{cat}])$
        \FOR{\textit{$x_{syn}$} in \textit{$X_{syn}$} }
        
        \STATE $Predicted$ = $CLF$ ($x_{syn} [F_{cont}] $)
        \IF{$predicted$ != $x_{syn} [cat] $}
        \STATE discard $x_{syn}$ from $X_{syn}$
        \ENDIF
        \ENDFOR
        \ENSURE Verified $X_{syn}$
    \end{algorithmic}
\end{algorithm}

\subsection{Labeling the Synthetic Data}
\label{sect:labeling}
In the case of supervised machine learning, domain experts are generally asked to label the data.  
However, this process can be very expensive and time-consuming.  Furthermore, in many settings, 
researchers may not have access to domain experts.  Instead, we use decision rules extracted from a 
random forest model trained on real training data.

In a decision tree, each path starting from the root and ending at a leaf can be considered a 
\emph{rule}.  Hence, as shown in Fig.~\ref{fig:diagram}, we investigate labeling of synthetic
data with rules derived from the random forest model that performs the best on the validation
set.  We evaluate various random forest models using different splitting criteria (such as Gini 
index and entropy) and different depth constraints on its constituent decision trees. 

\subsection{TUTOR Training Schemes}
\label{sect:schemes}
In this section, we explain how TUTOR uses labeled synthetic data to obtain a prior on the 
DNN weights. We introduce two schemes, termed Schemes A and B, to combine the 
synthetic and real data to train two FC DNN models.

\subsubsection{Scheme A}
Scheme A combines gradient-based learning with the Hierarchical Bayes approach to estimate the 
parameters of the model in two steps (as explained in Section \ref{sect:gradient+bayes}). 
Algorithm \ref{alg:scheme-a} summarizes the process of training DNN weights using synthetic and 
real data.  Scheme A uses the data generated by the three different synthetic data generation methods 
to pre-train the network architecture.  Pre-training is followed by use of the real training dataset 
to fine-tune and train network weights.  Scheme A outputs the model that performs the best on the 
validation set.  Then, we use the test set for its evaluation.

\begin{algorithm}[h]
    \caption{Scheme A}
    \label{alg:scheme-a}
    \begin{algorithmic}[l]
        \REQUIRE \textit{($X$,$y$)}: Training data; \textit{Methods}: the three methods for synthetic 
data generation; \textit{Arch}: DNN architecture;         $KB$: knowledge-base (random forest model)
        \FOR{\textit{method} in \textit{Methods} }
        \STATE $X_{\text{Syn}}$: synthetic data based on \textit{method}
        \STATE $y_{syn}$ = KB ($x_{syn}$)
        \STATE Pre-train \textit{Arch} with ($X_{\text{Syn}}$, $y_{\text{Syn}}$)
        \STATE Train \textit{Arch} with ($X$, $y$)
        \STATE $Model_{method}$ = \textit{Arch} with the learned weights
        \ENDFOR
        \STATE $Model_{A}$: best model as evaluated on the validation set
        \STATE $testAcc$: Evaluate $Model_{A}$ on test data
        \ENSURE $Model_{A}$ and $testAcc$
    \end{algorithmic}
\end{algorithm}

\subsubsection{Scheme B}
Algorithm~\ref{alg:scheme-b} summarizes Scheme B.  First, we train two instances of the same DNN 
architecture.  We train the first DNN on the training dataset.  This network is responsible for 
learning the data.  We train the second DNN on synthetic data. This network is responsible for 
learning the KB, i.e., the random forest model.  Next, we obtain another architecture by concatenating 
the outputs of these two networks and adding two FC layers followed by an output layer. We then train 
this architecture on the training dataset and evaluate it on the validation set. 

The two added FC layers are responsible for learning the importance of the output from each of the 
two DNN components in making the final prediction.  In this aspect, Scheme B takes inspiration from 
an MLN \cite{richardson2006markov}.  An MLN associates a weight with each FOL formula in a KB to 
depict its relative importance.  Furthermore, it uses Markov networks to model uncertainty in 
reasoning based on the KB.  Similarly, Scheme B uses the added two FC layers to learn the relative 
importance of prediction based on training data and based on synthetic data that represent 
the KB, in the final prediction.  Finally, we record the best model and evaluate it on the test set.

\begin{algorithm}[h]
    \caption{Scheme B}
    \label{alg:scheme-b}
    \begin{algorithmic}[l]
        \REQUIRE \textit{($X$,$y$)}: Training data; \textit{Methods}: the three methods for synthetic 
data generation;  \textit{Arch}: DNN architecture; $KB$: knowledge-base (random forest model)
        \FOR{\textit{method} in \textit{Methods} }
        \STATE $X_{\text{Syn}}$: synthetic data based on \textit{method}
        \STATE $y_{syn}$ = KB ($x_{syn}$)
        \STATE $Model_1$: Train \textit{Arch} on ($X_{\text{Syn}}$, $y_{\text{Syn}}$)
        \STATE $Model_2$: Train \textit{Arch} using ($X$,$y$)
        \STATE \textit{CombinedNet}: \texttt{Concat} (Output of \textit{Models} $1$ and $2$) + $2$ FC 
layers + an output layer 
        \STATE fine-tune \textit{CombinedNet} on the training set
        \STATE Evaluate on the validation set
        \ENDFOR
        \STATE $Model_{B}$: best \textit{CombinedNet} on the validation set
        \STATE $testAcc$: Evaluate $Model_{B}$ on test data
        \ENSURE $Model_{B}$ and $testAcc$
    \end{algorithmic}
\end{algorithm}

\subsection{Grow-and-prune Synthesis}
\label{sect:gp}
In this section, we discuss the grow-and-prune synthesis step.
This approach, first proposed in \cite{dai2019nest, hassantabar2019scann}, allows both the 
architecture and weights to be learned during the training process, enhancing both model accuracy and
compactness.  We apply grow-and-prune synthesis to models $A$ and $B$ that are outputs of 
Algorithms~\ref{alg:scheme-a} and \ref{alg:scheme-b}, respectively. 

The approach we use allows the depth of the DNN to change during the training process. 
This is enabled by allowing a neuron to feed its output to any neuron activated after it.
As a result, DNN depth depends on how the neurons are connected in the
architecture and can be changed during the grow-and-prune synthesis process. 
We use three different architecture-changing operations for a predefined number of 
iterations (set to five in our experiments). After each change, the DNN architecture is trained for 
a few epochs (set to $20$ in our experiments) and its performance evaluated on the validation set. 
Finally, we choose the highest-performing architecture on the validation set.

The three operations that are used in our grow-and-prune synthesis step are as follows:

\noindent
\textbf{Connection growth}: 
Connection growth activates the dormant connections in the network, with their weights set to $0$ 
initially. The two main approaches we use for connection growth are: 

\begin{itemize}
    \item \textbf{Gradient-based growth}: In this approach, we choose the added connections based on 
their effect on the loss function $\mathcal{L}$. Each weight matrix $W$ has a corresponding binary 
mask $Mask$ that has the same size.  The binary mask is used to disregard the inactive connections 
with zero-valued weights.  Algorithm \ref{alg:gradient-growth} shows the process of gradient-based 
growth.  The goal of this process is to identify the dormant connections that are most effective in 
reducing the loss function value.  For each mini-batch, we extract the gradients for all the weight 
matrices $W.grad$ and accumulate them over a training epoch.  An inactive connection is 
activated if its gradient is above a certain percentile threshold of the gradient magnitude of its 
associated layer matrix.
    \item \textbf{Full growth}: This approach restores all the dormant connections in the network to 
make the DNN fully connected.
\end{itemize}

\begin{algorithm}[h]
    \caption{Connection growth algorithm}
    \label{alg:gradient-growth}
    \begin{algorithmic}[l]
        \REQUIRE 
        $W \in R^{M \times N}$: weight matrix of dimension $M \times N$;
        $Mask \in R^{M \times N}$: weight mask of the same dimension as the weight matrix;
        Network $P$; $W.grad$: gradient of the weight matrix (of dimension $M \times N$); data $D$; 
        $\alpha$: growth ratio
        \IF{full growth}
        \STATE $Mask_{[1:M, 1:N]} = 1 $
        %\STATE {set all elements in $Mask$ to 1}
        %\ELSIF{random growth}
        %\STATE {randomly set some elements in $C$ to 1}
        \ELSIF{gradient-based growth}
        \STATE {Forward propagation of data $D$ through network $P$ and then back propagation}
        \STATE {Accumulation of $W.grad$ for one training epoch}
        \STATE {$t = (\alpha \times MN)^{th}$ largest element in the $\left|W.grad\right|$ matrix} 
        
        \FORALL {$w.grad_{ij}$}
        \IF{$\left| w.grad_{ij} \right| > t$}
        \STATE {$Mask_{ij} = 1$}
        %\STATE {$w_{ij} = 0$}
        \ENDIF
        \ENDFOR
        %\FORALL {$w_{ij}$}
        %\IF{$\left| w.grad_{ij} \right| < t$}
        %\STATE {$Mask_{ij} = 0$}
        %\STATE {forward propagation through $N$ using data $D$ and then back propagation}
        %\STATE {compute $g_{ij} = \left|\frac{\partial L}{\partial u_j}x_i\right|$}
        %\STATE {For $g_{ij} > t$, set $c_{ij} = 1, w_{ij}=0 $ }
        \ENDIF
        \STATE $W$ = $W \otimes Mask$ 
        \ENSURE Modified weight matrix $W$ and mask matrix $Mask$
    \end{algorithmic}
\end{algorithm}

\noindent
\textbf{Connection pruning}: The connection pruning operation is based on the magnitude of the 
connection. 
We remove a connection $w$ if and only if its magnitude is smaller than a certain 
percentile threshold of the weight magnitude of its associated layer matrix. 
Algorithm \ref{alg:pruning} shows this process.

\begin{algorithm}[h]
    \caption{Connection pruning algorithm}
    \label{alg:pruning}
    \begin{algorithmic}[l]
        \REQUIRE Weight matrix $W \in R^{M \times N}$; mask matrix $Mask$ of the same dimension as 
        the weight matrix; $\alpha$: pruning ratio
        \STATE $t = (\alpha \times MN) ^{th}$ largest element in $\left|W\right|$
        \FORALL {$w_{ij}$}
        \IF{$\left| w_{ij} \right| < t$}
        \STATE {$Mask_{ij} = 0$}
        \ENDIF
        \ENDFOR
        \STATE $W$ = $W \otimes Mask$ 
        \ENSURE Modified weight matrix $W$ and mask matrix $Mask$
    \end{algorithmic}
\end{algorithm}

\noindent
\textbf{Neuron growth}: 
This step duplicates the existing neurons in the architecture, to add neurons to the network, and 
increases network size.  Algorithm \ref{alg:neuron-growth} explains the neuron growth process.
In this process, we select the neuron with the highest activation function (neuron $i$) for duplication. 
After 
adding the new neuron (neuron $j$) to the network, we use the original neuron $i$ to set the new values for mask and weight matrices.
In addition, to break the symmetry, random noise is added to the weights of all the 
connections related to the newly added neuron. 

\begin{algorithm}[h]
    \caption{Neuron growth algorithm}
    \label{alg:neuron-growth}
    \begin{algorithmic}[l]
        \REQUIRE Network $P$; weight matrix $W \in R^{M \times N}$; mask matrix $Mask$ of the same 
        dimension as the weight matrix; data $D$; candidate neuron $n_j$ to be added; array $A$ of activation values for all hidden neurons 
        %\IF{neuron division}
        %\IF{activation-based selection}
        \STATE {forward propagation through $P$ using data $D$}
        \STATE {$i = argmax~(A)$}
        %\ELSIF{random selection}
        %\STATE {randomly pick an active neuron $n_i$}
        %\ENDIF
        \STATE {$Mask_{[j, 1:N]} = Mask_{[i, 1:N]}$}
        \STATE {$Mask_{[1:M,j]} = Mask_{[1:M,i]}$}
        \STATE {$W_{[j,1:N]} = W_{[i,1:N]} + noise$} 
        \STATE {$W_{[1:M,j]} = W_{[1:M,i]} + noise$}
        
        \ENSURE Modified weight matrix $W$ and mask matrix $Mask$
    \end{algorithmic}
\end{algorithm}

We apply connection pruning after neuron growth and connection growth in each iteration. 
We apply these three operations for a pre-defined number of iterations. Finally, we select the 
best-performing architecture on the validation set.

\subsection{TUTOR Application: Using Synthetic Data Alone}
It is common to assume that a trusted curator gathers a dataset with sensitive information from a large 
number of individuals. Such a dataset can be used to train predictive machine learning models for 
specific tasks in a related domain.  A dataset can contain \emph{micro-data}, revealing information 
about individuals from whom the data are collected.  For example, many healthcare datasets contain 
legally protected information, such as health histories of patients. Hence, there is a privacy risk 
in publishing or sharing such datasets.  Narayanan et al.~\cite{narayanan2008robust} have shown 
that an adversary with a small amount of background knowledge about an individual can use it to 
identify the individual's record in an anonymized dataset, with a high probability. 
This means that the adversary can learn private information about the individual. 

In addition, a study \cite{sweeney2000simple} shows that using $1990$ U.S. Census summary data, with 
a high probability, $87\%$ of the U.S. population can be identified with only three features: 
ZIP Code, gender, and date of birth. 

As a result, the free flow of data between various organizations risks privacy loss.  However, data 
sharing is necessary for building predictive machine learning models.  To address this issue, we 
could share synthetic data, drawn from the same distribution as the real data, with the other parties 
\cite{surendra2017review}.  The synthetic data generation module in TUTOR can be used to generate such 
a dataset.  Furthermore, the dataset can be labeled using the most accurate DNN model (on the 
validation set) synthesized by the TUTOR framework.  In Section~\ref{sect:data-privacy}, we show that 
DNN models trained on the synthetic dataset incur little to no drop in accuracy compared to DNNs 
trained on real data. 

\section{Experimental Results}
\label{sect:results}
In this section, we evaluate the performance of the TUTOR synthesis methodology on nine datasets. 
Table \ref{tab:characteristics} shows their characteristics. These datasets are publicly available 
from the UCI machine learning repository \cite{ucidb}, Kaggle classification datasets 
\cite{ptbdb, mitbih, covertype}, and Statlog collection \cite{Hsu2002comparison}.  
Since TUTOR focuses on tabular datasets, none of these datasets are image-based.  In addition, TUTOR 
focuses on settings where limited data are available. Hence, the majority of the chosen datasets are 
of small to medium size.  First, we present results of applying TUTOR DNN synthesis to these datasets. 
In this case, we use the available data to synthesize the most accurate DNN models. We also evaluate 
the contribution of each training step of the TUTOR framework.  Next, we evaluate the ability of 
TUTOR to reduce the need for large amounts of data.  In this context, we evaluate the performance of 
TUTOR-generated models when only a part of the training and validation datasets are available.
We then look into using dimensionality reduction (DR) as a method to reduce the number of parameters 
in the joint density estimation, and generating higher quality synthetic data. 
Finally, we discuss an application of TUTOR in which only the synthetic data are shared
with other parties, not the real data.  We compare DNN models synthesized for each case.

\begin{table*}[htbp]
\caption{Characteristics of the datasets.}
\label{tab:characteristics}
\centering
\begin{tabular}{lccccc}
\toprule
Dataset  & Training Set & Validation Set & Test Set & Features & Classes \\ \midrule
Forest Cover Type (CoverType) \cite{covertype} & $348600$ & $116201$ & $116201$ & $12$ & $7$ \\ 
Sensorless Drive Diagnosis  (SenDrive)   \cite{ucidb}      & $40509$      & $9000$        & $9000$   & $48$  & $11$     \\
MIT-BIH Arrhythmia Dataset (BIH-arrhythmia)  \cite{mitbih} &  $22628$  & $7543$  & $7543$ & $187$  & $2$\\
PTB Diagnostic ECG Database (PTB-ECG) \cite{ptbdb} &  $8730$ & $2910$ & $2910$ & $187$ & $2$\\
Smartphone-based Human Activity Recognition (SHAR) \cite{ucidb}  & $6213$ & $1554$ & $3162$ & $561$ & $12$ \\

Epileptic Seizure Dataset (EpiSeizure) \cite{ucidb} &  $6560$  & $1620$  & $3320$ & $178$ & $2$\\
Human Activity Recognition Dataset (HAR)    \cite{ucidb}  & $5881$      & $1471$       & $2947$   & $561$   & $6$ \\
Statlog DNA (DNA) \cite{Hsu2002comparison} & $1400$ & $600$ & $1186$ & $180$ & $3$ \\
Breast Cancer Wisconsin Diagnostic Dataset (Breast Cancer) \cite{ucidb} & $404$      & $150$        & $160$   & $30$   & $2$     \\ \bottomrule 
%HAR - Reduced & $1500$ ($3.92\times$) &  $500$ ($2.94\times$) & $2947$ & $40$ ($14.02\times$) & $6$ \\ 

%BIH - Reduced &  $2000$ ($11.31\times$) & $240$  ($31.43\times$)& $7543$ & $69$ ($2.71\times$) & $2$\\

%PTB - Reduced &  $1680$ ($5.20\times$) & $560$ ($5.20\times$) & $2910$ & $50$ ($3.74\times$) & $2$ \\ \bottomrule 
\end{tabular}
\end{table*}

\subsection{TUTOR Performance Evaluation}
\label{sect:performance}

\begin{figure*}[!ht]
    \centering
    \includegraphics[scale=0.24]{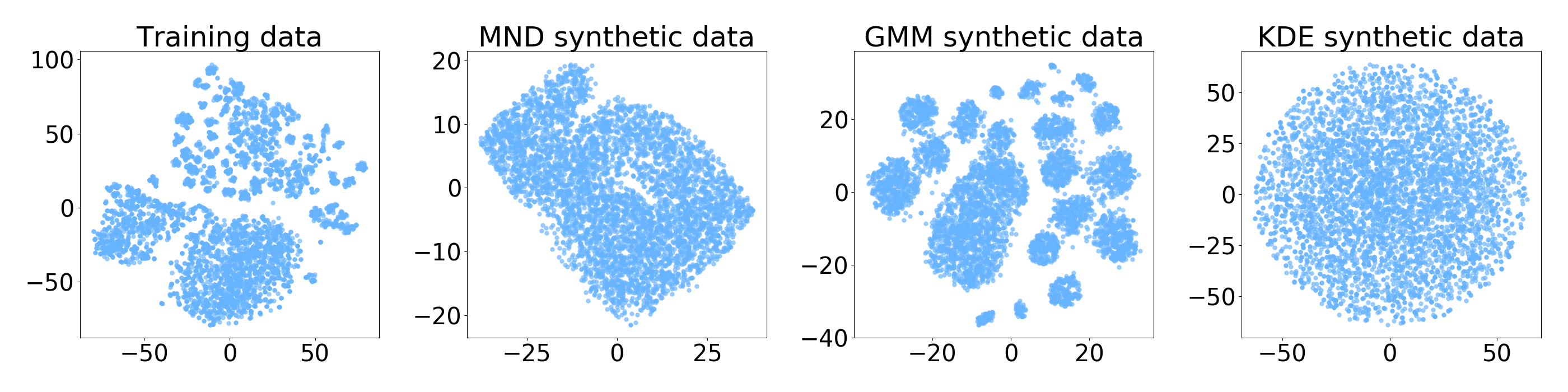}
    \caption{Visualizing data distribution using t-SNE for $5000$ data points obtained from
various data generation methods applied to the HAR dataset, from left to right: (1) original
training data, (2) multi-variate normal distribution, (3) GMM, and (4) kernel density estimation.}
\label{fig:tsne}
\end{figure*}

In this section, we evaluate the performance of the TUTOR framework on nine datasets.  
We present results for various synthetic data generation methods and training schemes. 
To compare the different synthetic data generation methods, we use t-distributed Stochastic Neighbor 
Embedding (t-SNE) \cite{maaten2008visualizing} to visualize high-dimensional data in a two-dimensional 
plane.  Fig.~\ref{fig:tsne} compares the distribution of the three synthetic data generation methods 
with that of the original training data for the HAR dataset.

\begin{figure}[!h]
    %\centering
    \includegraphics[scale=0.4]{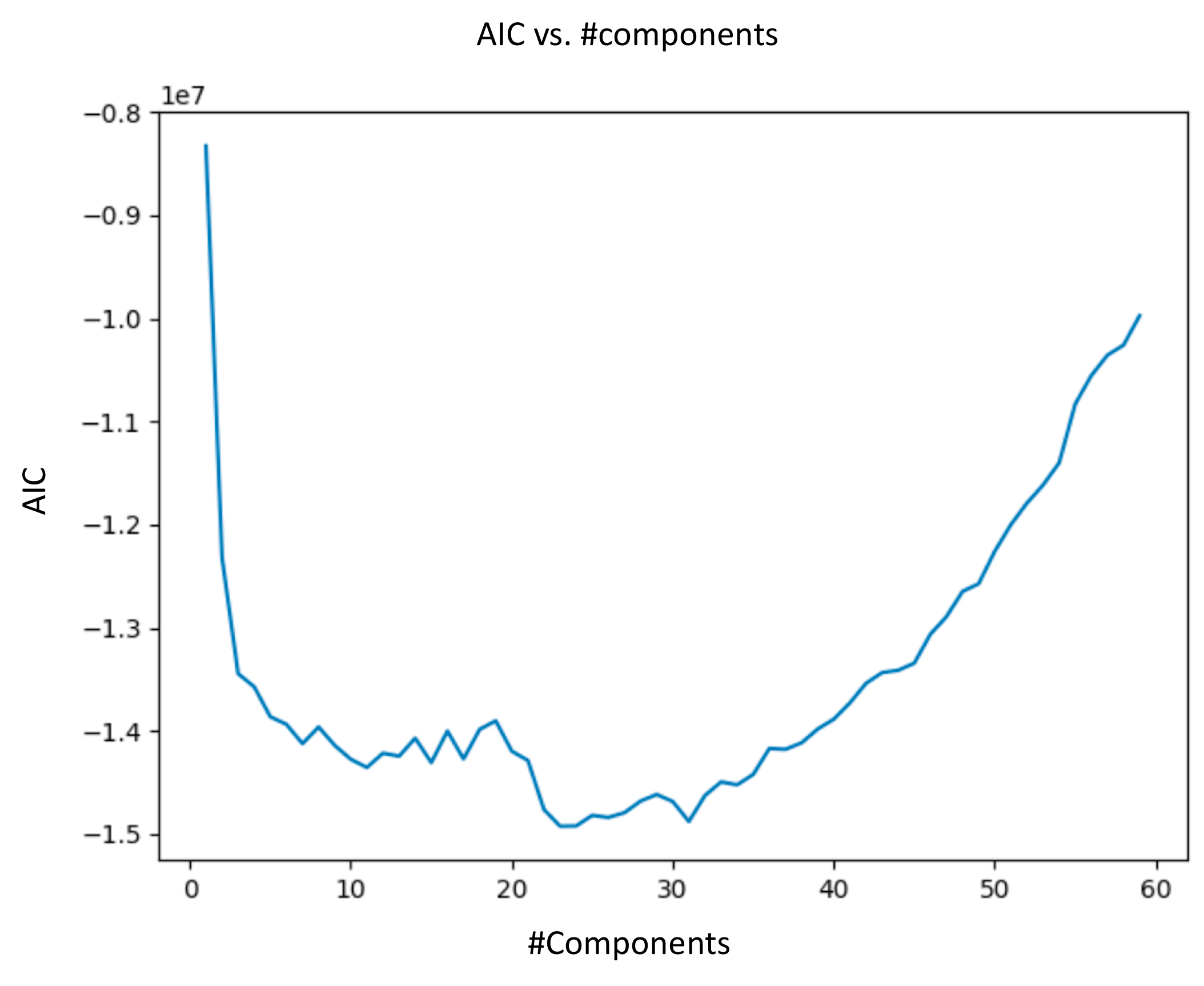}
    \caption{AIC vs. number of components in GMM for the HAR dataset.}
    \label{fig:aic}
\end{figure}

One of the methods used in statistics to compare various models and measure how well they fit the 
data is the Akaike information criterion (AIC) \cite{vrieze2012model}.  When a model is used to 
represent the process that generated the data, some of the information is lost due to the model not 
being exact.  AIC measures the relative information lost in the process of modeling that data.  A 
lower AIC value shows a better fit.  
Fig.~\ref{fig:aic} shows the AIC values as a function of the number of components in the GMM for
modeling the training data of the HAR dataset. 
As can be seen, for this metric the best fit happens when the GMM has 22 components. 

Next, we analyze the impact of different parts of TUTOR on DNN model performance. 
We compare three classes of DNN models. The first class consists of conventional FC DNNs trained on
the real training dataset. We refer to such models as DNN 1. The second class of models, referred to
as DNN 2, consists of the better model between those obtained by Schemes A and B. The third
class of models, referred to as DNN 3, is based on grow-and-prune synthesis. The starting point for 
DNN 3 is the better one from those synthesized by Schemes A and B, as evaluated on the validation set.

Table~\ref{tab:NN-ablation} shows the comparisons. We report test accuracy, FLOPs, and number of 
parameters (\#Param) for each model. It also shows the test accuracy of the random forest model used in labeling the synthetic dataset. 
For each dataset, we train various FC DNNs (with different 
number of layers and neurons per layer) and verify their performance on the validation set.  The 
FC baseline (DNN 1) for each dataset is the one that performs the best on the validation set. 
Since Scheme A only changes the weights of the network using synthetic data, the network architecture 
remains the same. Therefore, the FLOPs and number of parameters in the network are the same as 
those of DNN 1 models.  Compared to DNN 1 models, Schemes A and B that take help of synthetic data 
can be seen to perform better. On average, DNN 2 models have $2.6\%$ higher test accuracy
relative to DNN 1 models. This shows the importance of making use of synthetic data when the 
available dataset size is not large.  We summarize below the relative performance of the DNN models 
obtained through grow-and-prune synthesis using both real and synthetic data (DNN 3). 

\begin{itemize}
    \item \textbf{Better test accuracy:} Compared to DNN 1 (DNN 2), DNN 3 improves test accuracy 
    on an average by 3.4\% (2.3\%).
    \item \textbf{Less computation:} There is a $4.3\times$ ($5.9\times$) reduction in FLOPs per 
    inference on an average (geometric mean) relative to DNN 1 (DNN 2).
    \item \textbf{Smaller model size:} There is a $4.7\times$ ($6.3\times$) reduction in 
    the number of parameters on an average (geometric mean) relative to DNN 1 (DNN 2). Hence, the memory 
    requirements are also significantly reduced.
\end{itemize}

\small\addtolength{\tabcolsep}{-2.0pt}
\begin{table*}[]
\caption{Comparison results for various parts of TUTOR. The best test accuracies between
Schemes A and B (DNN 2) and with grow-and-prune (GP) synthesis are shown in bold. The results for the 
random forest (RF) model are also shown.}
\label{tab:NN-ablation}
\centering
\begin{tabular}{l|c|ccc|ccc|ccccc|ccc}
\toprule
             & \multicolumn{1}{c|}{RF} & \multicolumn{3}{c|}{FC Baseline}  & \multicolumn{3}{c|}{Scheme A (ACC. \%)} &
\multicolumn{5}{c|}{Scheme B (Acc. \%)} & \multicolumn{3}{c}{ + GP synthesis} \\
Dataset & Acc. (\%) & Acc. (\%) & FLOPs & \#Param. & MND  & GMM & KDE &  MND  & GMM  & KDE & FLOPs & \#Param. & Acc. (\%) & FLOPs & \#Param. \\
\toprule
CoverType & 91.5 &  91.3 & 90.0k & 45.2k & \textbf{92.6} & 91.3 & 92.6 & 92.4 & 91.8 &  92.5 & 184.2k & 92.5k & \textbf{93.1} & 19.5k & 10.0k \\
SenDrive & 96.6 & 95.1 & 19.3k & 9.0k & 96.5 & 97.3 & 95.9 & 97.4 & \textbf{97.5} & 97.1 & 44.0k & 20.8k & \textbf{99.0} & 9.8k & 5.0k \\
BIH-Arrhythmia & 95.3  &  94.5 & 107.4k & 54.1k & 96.0 & \textbf{96.5} & 96.1 & 94.9 & 95.5 & 94.9 & 215.1k & 108.4k & \textbf{96.6} & 9.7k & 5.0k \\
PTB-ECG &  95.9 & 95.0 & 21.2k & 10.7k & 95.2 & \textbf{96.2} & 95.6 & 95.8 & 95.1 & 95.8 & 42.7k & 21.6k & \textbf{97.3} & 19.9k & 5.0k \\
SHAR & 90.3 & 91.6 & 707.6k & 354.9k & 93.4 & 93.0 & 93.3 & 93.2 & \textbf{93.7} & 92.7 & 1.4M & 710.6k & \textbf{94.3} & 19.2k & 10.0k \\
EpiSeizure & 97.4 & 89.4 & 95.5k & 48.1k & 93.0 & 92.4 & 93.0 & \textbf{95.6} & 94.6 & 95.0 & 191.9k & 96.7k & \textbf{97.2} & 69.7k & 35.0k \\
HAR &  87.5 &  93.2 & 703.1k & 352.7k & \textbf{94.2} & 93.8 & 94.1 & 93.3 & 93.6 & 93.6 & 1.4M & 706.1k & \textbf{95.1} & 49.2k & 25.0k \\
DNA &  92.4 & 92.3 & 130.3k & 65.7k & 93.9 & 94.8 & 93.5 & 93.5 & \textbf{94.9} & 94.0 & 260.9k & 131.6k & \textbf{95.8} & 49.6k & 25.0k \\
Breast Cancer & 93.7  & 93.7 & 7.2k & 3.7k & 94.4 & \textbf{96.9} & 94.4 & 94.4 & 96.6 & 95.0 & 14.7k & 7.6k & \textbf{98.7} & 2.9k & 1.5k\\
\bottomrule
\end{tabular}
\end{table*}

\subsection{Reducing the Need for Data}
\label{sect:data-reduction}
\begin{figure*}[!ht]
    \centering
    \includegraphics[width=1.0\textwidth]{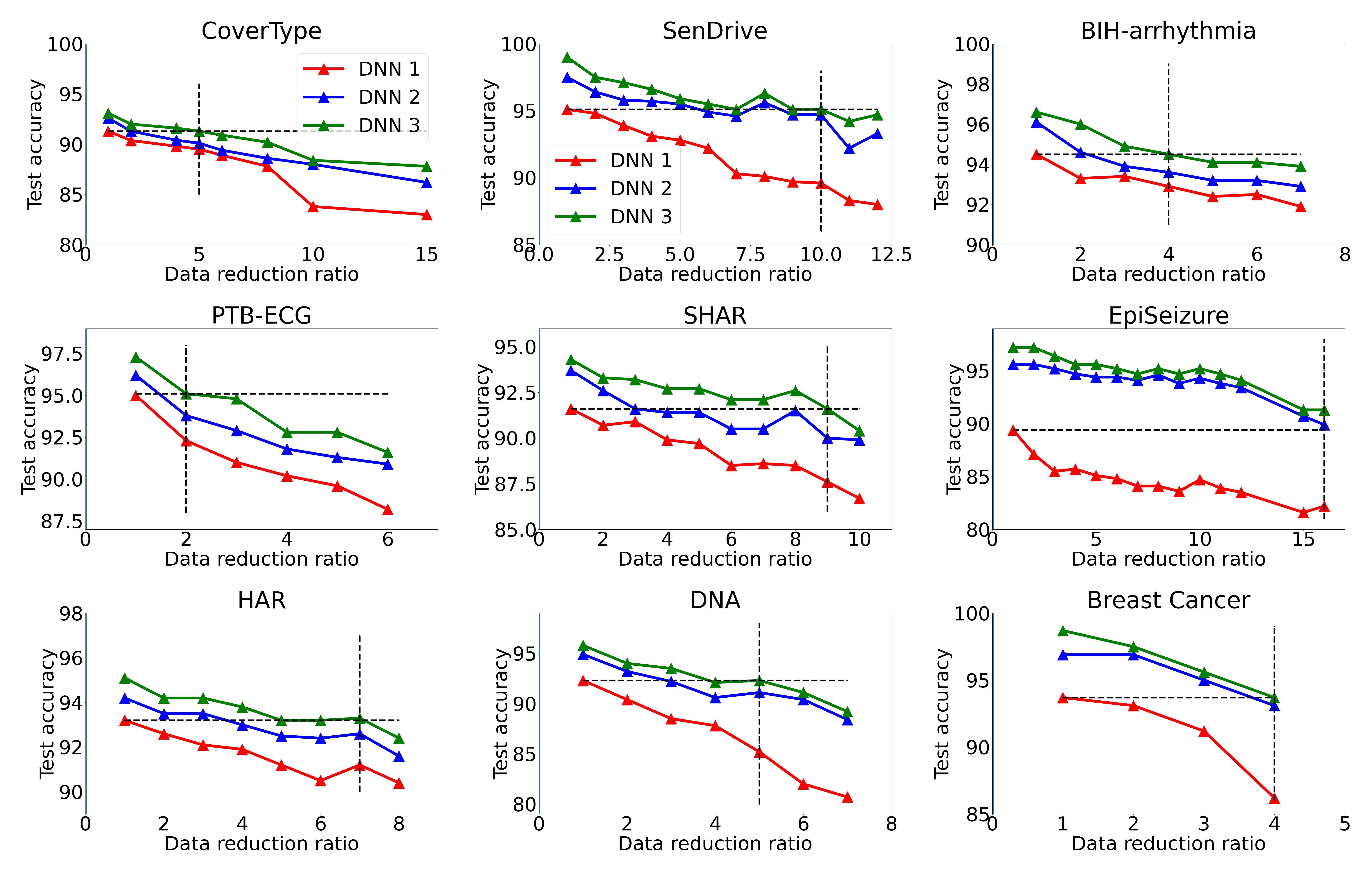}
    \caption{Test accuracy vs. data compression ratio.}
\label{fig:data-compression}
\end{figure*}

In this section, we explore the impact of TUTOR on reducing the number of data instances needed.
We apply TUTOR to only a part of the data selected using random sampling from the original dataset.
We define the \emph{data compression ratio} as the ratio of the original training (validation)
dataset size over the sub-sampled training (validation) dataset size.  For a fair comparison, the test 
set remains unchanged across various data compression ratios.  Fig.~\ref{fig:data-compression} shows 
the results for all the datasets. We compare the three DNN models as the dataset size decreases. 
The dashed horizontal and vertical lines in Fig.~\ref{fig:data-compression} show the needed data for 
DNN 3 to achieve similar test accuracy as DNN 1 that relies on the whole training and validation 
sets.  We summarize this information in Table~\ref{tab:reduced-data}.
On an average (geometric mean), TUTOR-synthesized DNN 3 (DNN 2) reduces the need for data by 
$5.9 \times$ ($3.6 \times$) while maintaining a similar test accuracy.

\begin{table*}[]
\caption{Comparison of the data needed for TUTOR-synthesized DNNs to achieve similar accuracy to DNN 1.}
\label{tab:reduced-data}
\centering
\begin{tabular}{l|ccc|ccc|ccc}
\toprule
             & \multicolumn{3}{c|}{DNN 1}  & \multicolumn{3}{c|}{DNN 2} & 
             \multicolumn{3}{c}{DNN 3} \\
Dataset & Acc. (\%) & \#Train & \#Val & Acc. (\%) & \#Train & \#Val &  Acc. (\%) & \#Train & \#Val\\
\midrule
CoverType & 91.3 & 348600 ($1\times$) & 116201 ($1\times$) & 91.3 & 174300 ($2\times$) & 58100 ($2\times$) &  91.3 & 69720 ($5\times$) & 23240 ($5\times$) \\
SenDrive & 95.1 & 40509 ($1\times$) & 9000 ($1\times$) & 94.6 & 5063 ($8\times$) & 1125 ($8\times$) & 95.1 & 4050 ($10\times$) & 900 ($10\times$) \\
BIH-Arrhythmia & 94.5 & 22628 ($1\times$) & 7543 ($1\times$) & 95.6 & 11314 ($2\times$) & 3771 ($2\times$) &  94.5 & 5657 ($4\times$) & 1885 ($4\times$) \\
PTB-ECG & 95.0 & 8730 ($1\times$) & 2910 ($1\times$) & 96.2  & 8730 ($1\times$) & 2910 ($1\times$) & 95.1 & 4365 ($2\times$) & 1455 ($2\times$) \\
SHAR & 91.6 & 6213 ($1\times$) &  1554 ($1\times$)  & 91.5 & 776 ($8\times$) & 194 ($8\times$)  & 91.6 & 690 ($9\times$) & 172 ($9\times$) \\
EpiSeizure & 89.4 & 6560 ($1\times$) & 1620 ($1\times$) & 89.9 & 410 ($16\times$) & 101 ($16\times$)   & 91.3 & 410 ($16\times$) & 101 ($16\times$)\\
HAR  & 93.2 & 5881 ($1\times$) & 1471 ($1\times$) & 93.5 & 1960 ($3\times$) & 490 ($3\times$)   & 93.3 & 840 ($7\times$) & 210 ($7\times$) \\
DNA & 92.3 & 1400 ($1\times$) & 600 ($1\times$) & 92.2 & 466 ($3\times$) & 200 ($3\times$) &  92.3 & 280 ($5\times$) & 120 ($5\times$) \\
Breast Cancer & 93.7 & 404 ($1\times$) & 150 ($1\times$) & 95.0 & 134 ($3\times$) & 50 ($3\times$) & 93.7 & 101 ($4\times$) & 38 ($4\times$) \\
\bottomrule
\end{tabular}
\end{table*}

We also compare the performance of TUTOR-generated synthetic data with the synthetic data generated 
by the CTGAN \cite{xu2019modeling} method.  In these experiments, we limit the available data to the 
same amount of data that DNN 2 and DNN 3 need to match the accuracy of DNN 1.  To this end, we compare 
the performance of Schemes A and B using the two synthetic datasets, one generated by TUTOR and the 
other generated by CTGAN.  Table~\ref{tab:comp-ctgan} summarizes the results.  For each case, we report 
the accuracy of DNN 2.  As can be seen, although the synthetic data generated by CTGAN performs better 
in a few cases, in a majority of cases our synthetic data leads to better results.  In addition, for 
the cases where the available data is smaller, the performance of the TUTOR is noticeably superior.
This is mainly due to the fact that GAN-based synthetic data generation relies on more data to train 
the generator and discriminator components. 

\begin{table*}[]
\caption{Comparison between the performance of synthetic data generated by our framework and CTGAN \cite{xu2019modeling}.}
\label{tab:comp-ctgan}
\centering
\begin{tabular}{l|cccc|cccc}
\toprule
             & \multicolumn{4}{c|}{DNN 2}  & \multicolumn{4}{c}{DNN 2} \\
Dataset &  Acc. (\%) & Acc. (\%) & \#Train & \#Val &  Acc. (\%) & Acc. (\%) & \#Train & \#Val\\
 & (CTGAN \cite{xu2019modeling}) & (Ours) &  &  & (CTGAN \cite{xu2019modeling}) & (Ours) & & \\
\midrule
CoverType & 91.5 & 91.3 & 174300 ($2\times$) & 58100 ($2\times$) &  89.7 & 90.1 & 69720 ($5\times$) & 23240 ($5\times$) \\
SenDrive &  93.6 & 94.6 & 5063 ($8\times$) & 1125 ($8\times$) & 92.9 & 95.1 & 4050 ($10\times$) & 900 ($10\times$) \\
BIH-Arrhythmia &  93.9 & 95.6 & 11314 ($2\times$) & 3771 ($2\times$) &  92.6 & 93.6  & 5657 ($4\times$) & 1885 ($4\times$) \\
PTB-ECG &  95.1  & 96.2 & 8730 ($1\times$) & 2910 ($1\times$) & 94.1 & 93.8  & 4365 ($2\times$) & 1455 ($2\times$) \\
SHAR &  87.8 & 91.5 & 776 ($8\times$) & 194 ($8\times$)  & 85.9 & 91.6 & 690 ($9\times$) & 172 ($9\times$) \\
EpiSeizure &  86.7 & 89.9 & 410 ($16\times$) & 101 ($16\times$) & - & - & - & - \\
HAR  &  92.0 & 93.5  & 1960 ($3\times$) & 490 ($3\times$)  & 90.3 & 92.6 & 840 ($7\times$) & 210 ($7\times$) \\
DNA & 89.2 & 92.2 & 466 ($3\times$) & 200 ($3\times$) &  86.9 & 91.1 & 280 ($5\times$) & 120 ($5\times$) \\
Breast Cancer & 86.9 &  95.0 & 134 ($3\times$) & 50 ($3\times$) & 85.0 & 93.1 & 101 ($4\times$) & 38 ($4\times$) \\
\bottomrule
\end{tabular}
\end{table*}

\subsection{Combining DR with TUTOR}
In this section, we look into the impact of reducing the number of features in the dataset by applying 
DR methods.  As the number of features in the data increases, more data instances are needed to model 
the joint distribution of the features in the data.  This is due to the increase in the number of 
independent parameters of the joint density model.  Thus, reducing the number of features may be 
helpful in reducing the need for more data instances.  Hence, we study the impact of reducing the 
number of features for the two most high-dimensional datasets in our experiments: SHAR, and HAR. 
We use the traditional principal component analysis (PCA) for dimensionality reduction.  Dimensionality 
reduction is only applied to continuous features and the categorical features remain unchanged. 
Table~\ref{tab:DR} shows the results.  For each dataset, we report the best accuracy of Schemes A and 
B for various feature compression ratios.  As we can see, DR helps improve model performance. 
For the HAR dataset, the highest accuracy corresponds to $2\times$ feature compression ratio using 
Scheme B for DNN synthesis.  A $3\times$ feature compression ratio and Scheme B yield the highest 
accuracy for the SHAR dataset.  Note that after a certain point, DR does not help improve performance, 
as it leads to loss of information. 

\small\addtolength{\tabcolsep}{-1.5pt}
\begin{table*}[]
\caption{Impact of DR on the performance of the TUTOR training schemes.}
\label{tab:DR}
\centering
\begin{tabular}{c|ccc|ccc}
\toprule
             & \multicolumn{3}{c|}{HAR data}  & \multicolumn{3}{c}{SHAR data}  \\
Feature compression ratio & DNN 1  & Scheme A & Scheme B & DNN 1  & Scheme A  & Scheme B  \\
 & (\%) & (\%) & (\%) & (\%)  & (\%)  & (\%)  \\
\toprule
1$\times$ & 93.2 & 94.2 & 93.6 & 90.3 & 93.4 & 93.7 \\
2$\times$ & 93.3 & 94.1 & 95.2 & 93.5 & 94.3 & 94.1 \\
3$\times$ & 93.7 & 93.7 & 94.6 & 93.7 & 94.2 & 95.1 \\
4$\times$ & 92.7 & 92.9 & 93.1 & 93.1 & 93.7 & 94.1\\
5$\times$ & 92.6 & 93.1 & 93.3 & 92.4 & 93.3 & 93.4 \\
6$\times$ & 92.4 & 92.9 & 93.7 & 92.3 & 92.5 & 93.1 \\
7$\times$ & 91.8 & 92.3 & 92.7 & 91.5 & 91.6 & 92.9 \\
\bottomrule
\end{tabular}
\end{table*}

\subsection{TUTOR Application: Performance of Synthetic Data Alone}
\label{sect:data-privacy}
In this section, we evaluate the application of TUTOR to generating synthetic data that can be used 
alone to train DNN predictive models.  In this application, to avoid exchanging personal information 
between various parties, we share the synthetic data generated by TUTOR instead.  We use the 
synthetic data generation scheme that leads to the most accurate DNN model on the
validation set. We use this DNN model to label the synthetic dataset.  We generate 100,000 synthetic 
data instances.

We train a generic three-layer DNN architecture, with two hidden layers with 100 neurons and an output 
layer, with only the synthetic data, and evaluate its performance on the separate real test set. 
We compare the results obtained with DNN 1 that is trained using the real training and validation sets.
Table~\ref{tab:NN-synvreal} shows the results. As one can see, using synthetic data alone to train 
the DNN model leads to similar performance as DNN 1 on the unseen test data.  Since we can generate 
large synthetic datasets, in six of the nine cases, the test accuracy improves by 0.2\% to 7.2\%. 

\small\addtolength{\tabcolsep}{-0.5pt}
\begin{table*}[]
\caption{Comparison of the FC baseline trained on real data with an FC DNN trained on synthetic data.}
\label{tab:NN-synvreal}
\centering
\begin{tabular}{l|ccc|cccc}
\toprule
             & \multicolumn{3}{c|}{FC DNN trained on real data}  & \multicolumn{4}{c}{FC DNN trained on syn. data}  \\
Dataset & Acc. (\%) & FLOPs & \#Param. & Method & Acc. (\%)  & FLOPs  & \#Param.  \\
\toprule
CoverType & 91.3 & 90.0k & 45.2k & GMM & 84.2 & 24.2k & 12.2k  \\
SenDrive & 95.1 & 19.3k & 9.0k & GMM & 92.5 & 31.6k & 16.1k  \\
BIH-Arrhythmia & 94.5 & 107.4k & 54.1k & KDE & 95.1 & 57.6k & 29.1k \\
PTB-ECG & 95.0 & 21.2k & 10.7k & GMM & 92.6 & 57.6k & 29.1k  \\
SHAR & 91.6 & 707.6k & 354.9k & KDE & 92.8 & 134.4k & 67.5k  \\
EpiSeizure & 89.4 & 95.5k & 48.1k & GMM & 96.6 & 55.8k & 28.2k \\
HAR & 93.2 & 703.1k & 352.7k & GMM & 93.4 & 133.2k & 66.9k  \\
DNA & 92.3 & 130.3k & 65.7k & GMM & 94.2 & 56.4k & 28.5k  \\
Breast Cancer & 93.7 & 7.2k & 3.7k & KDE & 95.0 & 26.2k & 13.4k \\
\bottomrule
\end{tabular}
\end{table*}
\section{Discussion and Limitations}
\label{sect:discussion}
In this section, we discuss the inspiration we took from the human brain in designing the TUTOR
framework. We also discuss TUTOR's limitations that can be addressed in the future. 

An interesting aspect of the human brain is its ability to quickly solve a problem in a new
domain, despite limited experience. It intelligently utilizes prior learning
experiences to adapt to the new domain. Inspired by this process, TUTOR addresses the need
for a large amount of data by generating synthetic data from the same probability
distribution as the limited available data. The synthetic data are used to warm-start the DNN
training process, providing it with an appropriate inductive bias. 

The human brain also changes its synaptic connections dynamically, to adapt to the task at hand. In
fact, this is one way for it to acquire knowledge. TUTOR takes inspiration from this aspect of human 
intelligence in the grow-and-prune synthesis step.  Allowing architectures to adapt during the 
training process enables TUTOR to generate accurate and efficient (both in terms of computation and 
memory requirements) architectures for the task at hand.

Although TUTOR addresses the small data problem of tabular datasets, the same problem may exist in 
image-based datasets.  For example, image-based datasets in healthcare tend to be
small, due to the difficulty of data collection and labeling.
Although techniques like DeepInversion \cite{yin2020dreaming} have begun to address realistic but
synthetic images from the same probability distribution, extending TUTOR to image-based
applications can also be part of future work.

\section{Conclusion}
\label{sect:conclusion}
In this work, we proposed the TUTOR framework to address the need for large datasets in training DNN 
predictive models.  TUTOR targets tabular datasets.  It relies on the synthetic data generated from the 
same distribution as the real training data.  Using the semantic integrity classifier module, TUTOR 
verifies the validity of the generated synthetic data. 
To label the verified synthetic data, it uses a random forest model trained on the real training data. 
Synthetic data are used alongside real data in two training schemes to train the DNN weights. 
These schemes employ synthetic data in two different ways to impose a prior on the DNN weights, thus 
starting the training process with a better initialization point.  
In addition, TUTOR utilizes the grow-and-prune synthesis paradigm to ensure model compactness while 
boosting accuracy.  TUTOR can be particularly useful in settings where available data are 
limited, such as healthcare applications.  Since the generated models are compact, they can be 
deployed on edge devices, such as smartphones and Internet-of-Things sensors. 

% Acknowledgment 
\bibliographystyle{IEEEtran}
\bibliography{Tutor}

\vspace{-1.0cm}
\begin{IEEEbiography}[
{\includegraphics[width=1.0in,height=1.1in,clip,keepaspectratio]{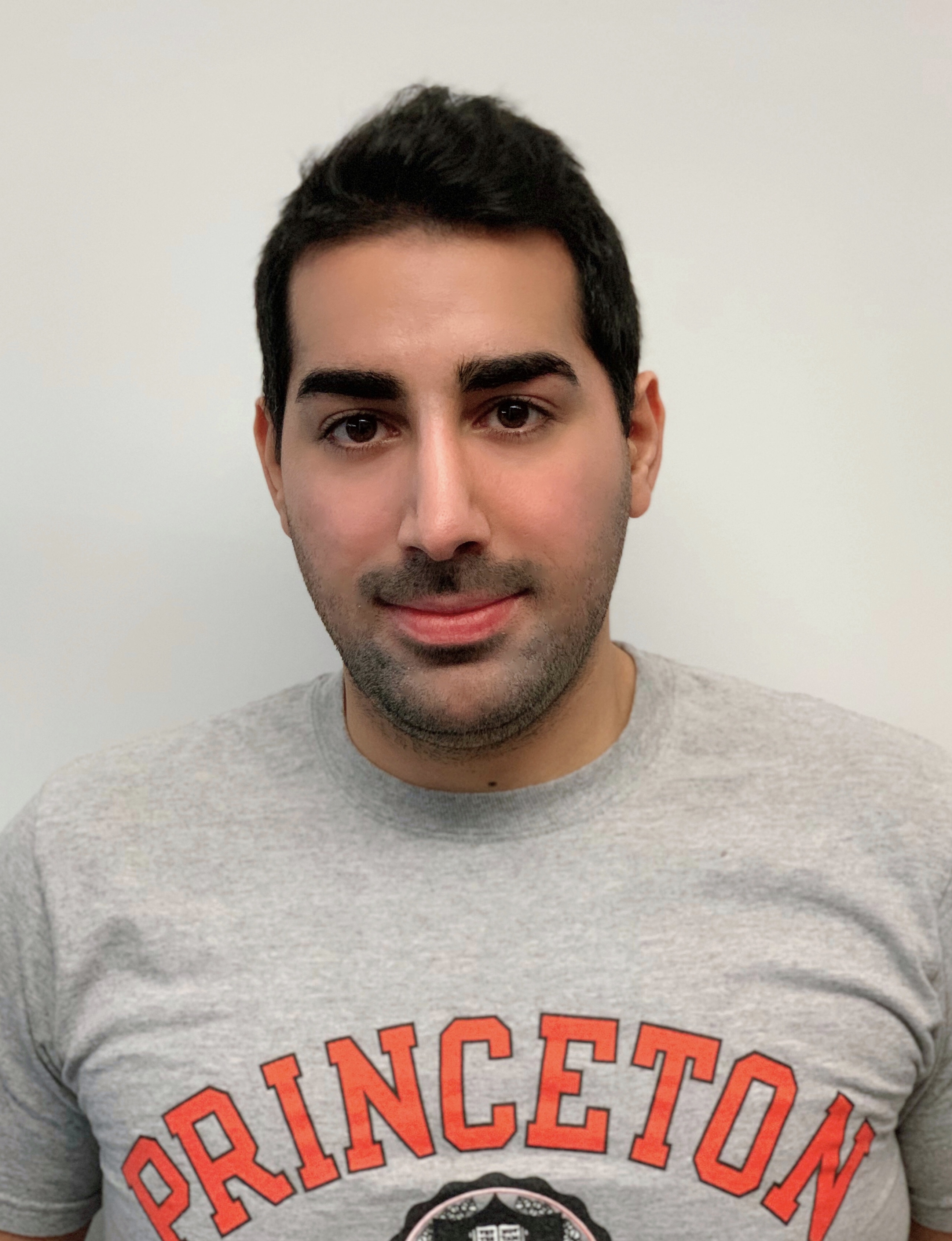}}]
{Shayan Hassantabar} received his B.S. degree in Electrical Engineering, Digital Systems focus, from Sharif University of Technology, Iran. He also received his M.Math. degree in Computer Science from University of Waterloo, Canada, and his M.A. degree in Electrical Engineering from Princeton University. He is pursuing the Ph.D. degree in Electrical Engineering at Princeton University. His research interests include automated neural network architecture synthesis, neural network compression, and smart healthcare.
\end{IEEEbiography}

\vspace{-1.0cm}
\begin{IEEEbiography}[{\includegraphics[scale = 0.35,keepaspectratio]{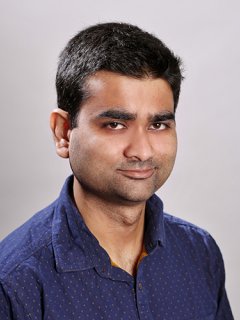}}]{Prerit Terway}
received his M.S. degree from University of Michigan, Ann Arbor, and B.Tech. degree from Indian Institute of Technology, Gandhinagar, India, both in Electrical Engineering. He is currently a Ph.D. candidate in Electrical Engineering at Princeton University. His research interests include machine learning and cyber-physical systems. 
\end{IEEEbiography}

\vspace{-1.0cm}
\begin{IEEEbiography}[
{\includegraphics[width=1.0in,height=1.1in,clip,keepaspectratio]{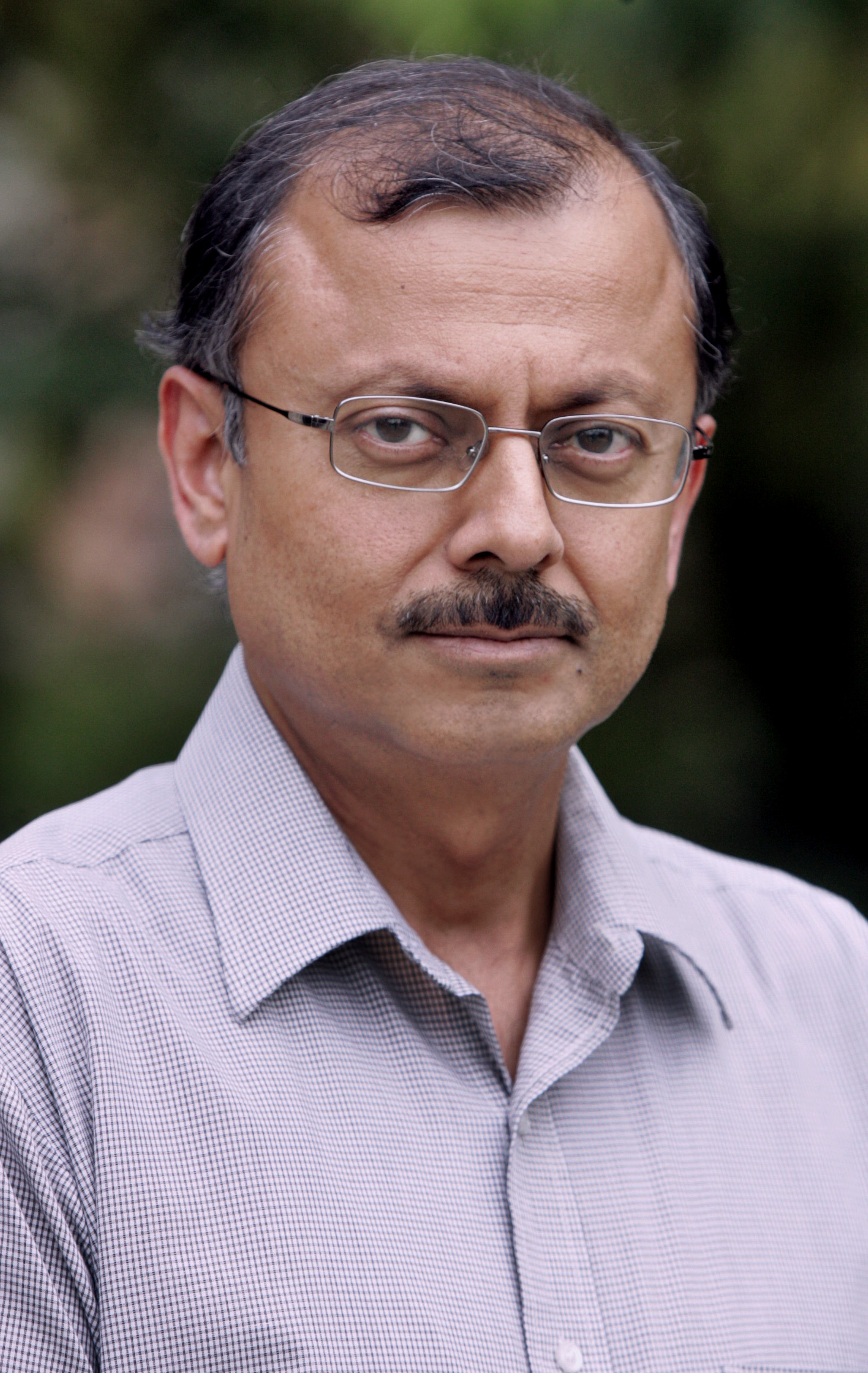}}]
{Niraj K. Jha} received his B.Tech. degree in Electronics and Electrical Communication Engineering from Indian Institute of Technology, Kharagpur, India in 1981 and Ph.D. degree in Electrical Engineering from University of Illinois at Urbana-Champaign, IL in 1985. He has been a faculty member of the Department of Electrical Engineering, Princeton University, since 1987. He is a Fellow of IEEE and ACM, and was given the Distinguished Alumnus Award by I.I.T., Kharagpur. He has also received the Princeton Graduate Mentoring Award.

He has served as the Editor-in-Chief of IEEE Transactions on VLSI Systems and as an Associate Editor of
several other journals. He has co-authored five books that are widely used, and 440 papers. His research has won 
20 best paper awards or nominations and 23 patents.  His research interests include smart healthcare, 
cybersecurity, machine learning, and CNN-accelerator co-design. He has given several keynote speeches in the 
area of nanoelectronic design/test, smart healthcare, and cybersecurity. 
\end{IEEEbiography}
\end{document}